\documentclass[10pt,twocolumn,letterpaper]{article}

\usepackage{cvpr}

\usepackage{graphicx}
\usepackage{amsmath}
\usepackage{amssymb}
\usepackage{booktabs}
\usepackage{multirow}
\usepackage{threeparttable}
\usepackage{listings}
\usepackage{float}
\usepackage{makecell}
\usepackage[export]{adjustbox}
\usepackage{changepage}
\usepackage{tcolorbox}
\usepackage{tikz}

\usepackage[pagebackref,breaklinks,colorlinks]{hyperref}

\usepackage[capitalize]{cleveref}
\crefname{section}{Sec.}{Secs.}
\Crefname{section}{Section}{Sections}
\Crefname{table}{Table}{Tables}
\crefname{table}{Tab.}{Tabs.}

\def\cvprPaperID{*****}
\def\confName{CVPR}
\def\confYear{2023}

\begin{document}

\title{H-POPE: Hierarchical Polling-based Probing Evaluation of Hallucinations in Large Vision-Language Models}

\author{
Nhi Pham$^*$\\
Max Planck Institute for Informatics\\
{\tt\small nhipham@mpi-inf.mpg.de}
\and
Michael Schott$^*$\\
Saarland University, Zuse School ELIZA$^{\dagger}$\\
{\tt\small misc00001@stud.uni-saarland.de}}
\maketitle
\def\thefootnote{*}\footnotetext{Authors contributed equally to this work, listed in alphabetical order.}
\def\thefootnote{$\dagger$}\footnotetext{\url{https://eliza.school/}}
\def\thefootnote{\arabic{footnote}}

\begin{abstract}
By leveraging both texts and images, large vision language models (LVLMs) have shown significant progress in various multi-modal tasks. Nevertheless, these models often suffer from hallucinations, \eg, they exhibit inconsistencies between the visual input and the textual output. To address this, we propose H-POPE, a coarse-to-fine-grained benchmark that systematically assesses hallucination in object existence and attributes. Our evaluation shows that models are prone to hallucinations on object existence, and even more so on fine-grained attributes. We further investigate whether these models rely on visual input to formulate the output texts.
\end{abstract}

\section{Introduction}

\label{sec:intro}
Recent advances in multi-modal models have enabled a wide array of impressive capabilities. For example, such models can generate a well-formulated description of a given image and thus reflect an understanding of both the textual and the visual domains. Despite their promising progress, existing models often suffer from a phenomenon called \emph{hallucination}, which not only degrades the model’s performance but also raises questions about safety and reliability. To understand and assess hallucinations in LVLMs, several benchmarks have been proposed. In specific, \textit{Caption Hallucination Assessment with Image Relevance} (CHAIR) \cite{rohrbachObjectHallucinationImage2019} demonstrates that when asked to provide an accurate description of a given image, responses from these models frequently include objects that are not actually present \cite{rohrbachObjectHallucinationImage2019}. Moreover, the \textit{Polling-based Object Probing Evaluation} (POPE) \cite{li2023evaluating} shows that these models tend to confirm the presence of an object that does not exist, if it frequently co-occurs with other objects in the image. 

While these aforementioned benchmarks effectively highlight the general issue of hallucination, they often focus on assessing hallucinations on object presence. A natural extension is to further assess hallucination on granular details such as attributes. Therefore, \emph{the main objective of our project is to conduct a more fine-grained evaluation on not only the existence of objects but also the attributes associated with them}. To achieve this goal, we leverage the discriminative approach outlined in POPE \cite{li2023evaluating}. For each image, we start with coarse-grained questions about the existence of objects and continue with fine-grained questions about the existence of attributes. The resulting benchmark is called \textit{Hierarchical} POPE (\textit{H-POPE}).

The original POPE includes three sampling strategies for negative objects. Some of them were specially designed to formulate challenging questions based on occurrence statistics in the dataset. We utilise these settings in our study, while also aiming to devise an additional sampling strategy for attributes. To this end, we propose \emph{Image-based Adversarial} setting, which shifts the focus to the local context of the given image and selects attributes that appear in the image but do not describe the object in question. Our evaluation shows that models tend to hallucinate worst on questions sampled in this setting, which suggests that existing LVLMs struggle to match attributes correctly to objects that possess them. 

Finally, it is noticeable that the current work centers their evaluation metrics around the models' textual output. We \emph{investigate relevance maps} to see if the models make different use of visual information when giving hallucinated answers versus correct answers.

In summary, our main contributions include:

\begin{itemize}
    \item Introducing H-POPE, a benchmark to assess hallucination on objects and attributes in LVLMs, including a new adversarial strategy targeted at evaluating whether LVLMs can match attributes to the correct objects in the image.
    \item Evaluating three current LVLMs, namely mPLUG-Owl \cite{mplugowl}, InstructBLIP \cite{instructblip} and LLaVa \cite{llava} on H-POPE.
    \item Qualitative examples of the visual cues used by LVLMs, which are derived from LVLM-Interpret \cite{stan2024lvlminterpretinterpretabilitytoollarge}.
\end{itemize}

\section{Related Work}
\subsection{Hallucination Benchmarks}
Hallucination evaluations of LVLMs, such as CHAIR \cite{rohrbachObjectHallucinationImage2019}, traditionally follow the generative paradigm. Given an instruction, models generate a textual description which will then be assessed for hallucination. However, this approach is often unstable due to its dependence on instruction design and caption length. To address this limitation, POPE introduces a novel discriminative framework for evaluating coarse-grained hallucinations \cite{li2023evaluating}. It assesses models' object hallucination with a set of polling binary questions based on objects present in the images and those that are taken from random, popular and adversarial negative sampling strategies. This has not only been proven to be stable across different prompts, but also allows for flexibility in adapting this benchmark for other types of hallucinations. 

\subsection{Large Scale Attributes (LSA) Dataset}
Most hallucination benchmarks leverage datasets with annotations for the evaluation. One such dataset is LSA, which combines attribute annotations extracted and aggregated from 6 different dataset: Visual Genome, GQA, Flickr30K-Entities, MS-COCO Captions 2017, COCO Attributes, and Localized Narratives \cite{pham2022improving}. LSA differs from existing works that only focus on adjective object attributes and ignore attributes of visual relationship among objects. Moreover, it takes advantage of negative label expansion: since many attributes are mutually exclusive, presence of such an attribute (\eg, clean) would automatically imply absence of other attributes (\eg, dirty) \cite{pham2022improving}. LSA fits well with the 
\subsection{Large Vision-Language Models (LVLMs)}
LVLMs leverage both textual and visual information to learn and generate content for different multi-modal tasks. These tasks include image captioning, where the model generates descriptive text based on a visual input, and image generation, where it produces visual content guided by textual prompts.  Several prominent examples of LVLMs include InstructBLIP \cite{instructblip}, LLaVa \cite{llava} and mPLUG-Owl \cite{mplugowl}. On a high level, these models consist of a pre-trained language encoder and a pre-trained vision encoder, which generate corresponding text and visual embeddings that are then aligned through various methods, enabling the model to understand and generate coherent multimodal outputs \cite{ghosh2024exploring}. Despite LVLMs' great promise in tackling various tasks, most models are still prone to hallucinations \cite{ghosh2024exploring}. In our study, we focus on investigating hallucinations in models that generate textual outputs given multi-modal input, namely an image and a textual prompt.

\section{Method}
\subsection{H-POPE Benchmark}
\begin{figure*}[h]
  \centering
  \includegraphics[width=0.9\textwidth]{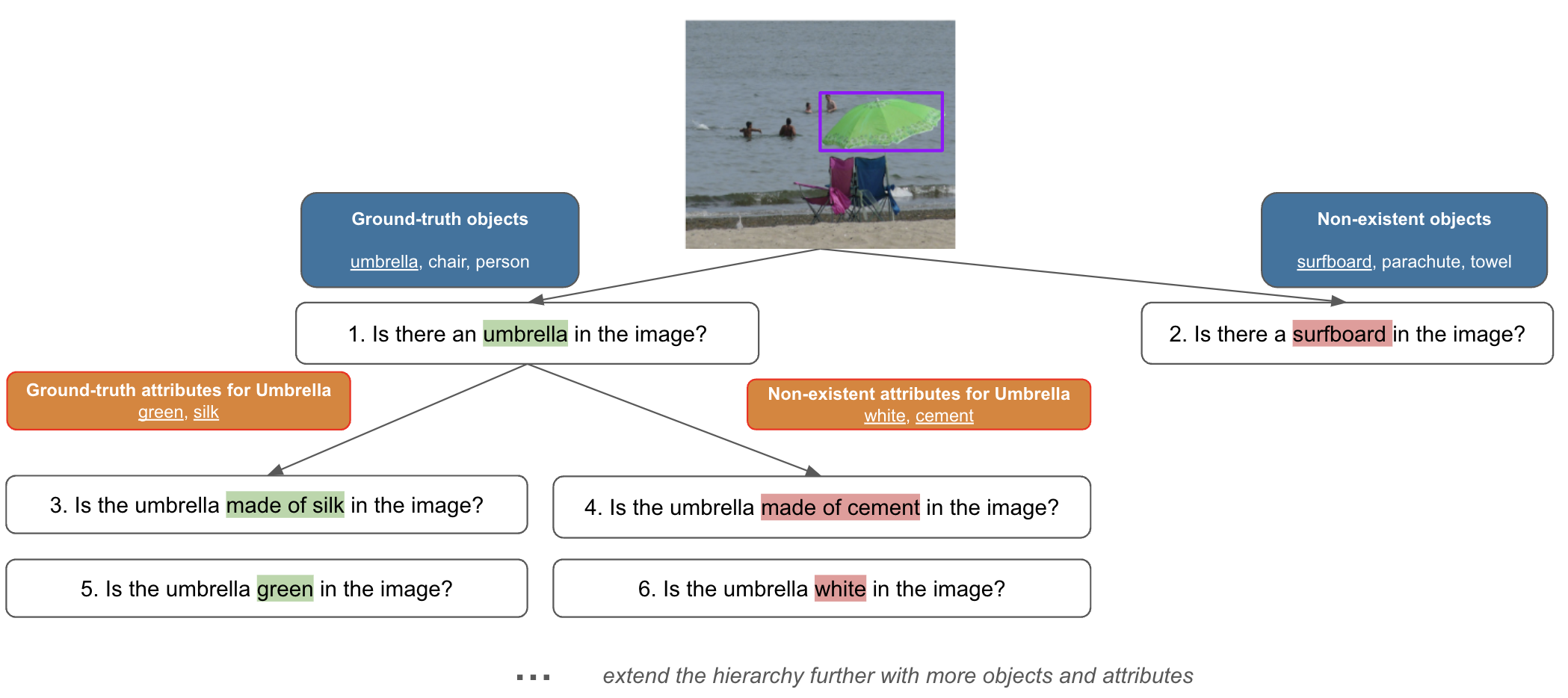}
  \caption{Overview of our H-POPE benchmark}
  \label{fig:hpopepipeline}
\end{figure*}

Our H-POPE benchmark extends POPE \cite{li2023evaluating} to include attribute evaluation. Given an input image, H-POPE extracts a list of ground-truth objects from the annotations, and samples an equal number of negative objects. These are used to formulate the first tier (coarser) of questions. Then, for each existent object, H-POPE extracts a list of ground-truth attributes and samples an equal number of negative attributes. These form the second tier (finer) of questions. Note that we sample objects and attributes with the same negative strategy. Our pipeline is captured in Figure \ref{fig:hpopepipeline}. 

\subsection{Negative Sampling Strategies}
\begin{figure}
     \centering
     \includegraphics[width=\linewidth]{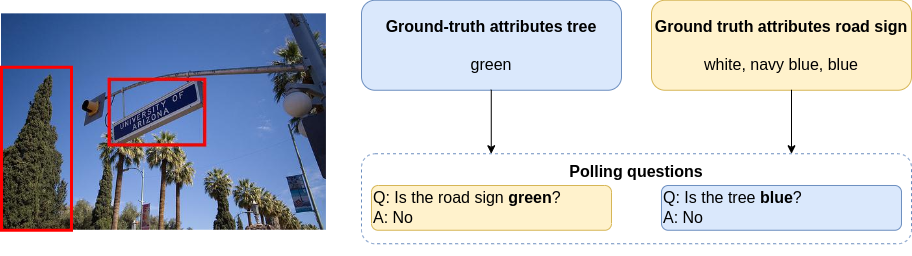}
     \caption{Overview of our image-based adversarial sampling strategy. For a given object, e.g.  the object \textit{road sign}, we randomly select an attribute from another object in the image, e.g. the attribute \textit{green} from the object \textit{tree}. We then ask whether the original object has that attribute.}
     \label{fig:adv-img}
 \end{figure}

For object existence, our negative sampling strategies follow those that are from the POPE benchmark \cite{li2023evaluating}. In specific:
 (i) \textbf{random}, which randomly samples any object from the dataset that does not exist in the image; (ii) \textbf{popular}, which selects top frequently appearing objects across the dataset that do not exist in the image; and (iii) \textbf{adversarial (frequency-based)}, which selects top co-occurring objects with the ground-truth objects that do not exist in the image. These three sampling strategies serve as a natural sampling scheme for attributes. We introduce an additional adversarial scenario --- (iv) \textbf{adversarial (image-based)} --- which selects attributes that are present in the image, because they describe other objects but do not describe the object in question. We illustrate this in \Cref{fig:adv-img}.

\subsection{Attribute Type Selection}
Our H-POPE benchmark formulates binary questions about object existence and object attributes. Negative attributes are sampled from the list of attributes across the dataset $\Omega$, so we need to ensure mutual exclusion, i.e., if an object has a list $A$ of attributes, it should not have any attributes from $\Omega \setminus A$. To this end, we restrict our setting to only three attribute types: \textit{color}, \textit{material}, and \textit{shape}. We use these types from the LSA dataset, because they are most frequent and typically mutually exclusive due to the negative label expansion in the dataset (e.g., if an object is annotated with \textit{red} and \textit{yellow}, it is likely that one can use other colors as negative attributes).

\section{Hallucination Evaluation}
\subsection{Evaluation Setting}
\noindent\textbf{Models.} We evaluate three popular LVLMs on our benchmark. These are InstructBLIP \cite{instructblip}, LLaVa \cite{llava} and mPLUG-Owl
\cite{mplugowl}. We choose the 7B variants for each of them. We query the models with an input image and the following prompts for object exsistence and attributes respectively:
\begin{footnotesize}
    \begin{verbatim}
Q: Is there a(n) <object> in the image?
Q: Is the <object> of <attribute> in the image?
\end{verbatim}
\end{footnotesize}

\noindent\textbf{Metrics.} Our benchmark is formulated as a binary
classification task, and thus we report accuracy, F1-score, precision and recall.

\noindent\textbf{Data.} Our evaluation uses images and objects from the MSCOCO (val. 2014)
\cite{lin2014microsoft}, and attribute annotations from the LSA dataset \cite{pham2022improving}. We select all images for which we have at least six questions (two questions on object existence, and four questions on attributes). For images for which we have more than six questions, we sample six questions at random. Additionally, if an image has multiple annotated objects, we select a random one to prevent some images from appearing more frequently than others. With this we are left with 994 images for random-, popular and frequency-based adversarial sampling and 926 images for our image-based sampling strategy. The exact numbers are shown in \Cref{tab:nimages}.
\begin{table}[h]
\centering
    \begin{tabular}{lr}
    \toprule
    Sampling Setting & \#Images \\
    \midrule
    Random & 994 \\
    Popular & 994 \\
    Frequency-based Adversarial & 994 \\
    Image-based Adversarial & 926 \\
    \bottomrule
    \end{tabular}
    \caption{Number of unique images used in our analysis for each sampling strategy.}
    \label{tab:nimages}
\end{table}

\subsection{Evaluation Results}

Our results are summarised in \Cref{tab:overall}, which includes the overall performance as well as a breakdown by object presence and attributes. The highest accuracy achieved by any model across all settings is roughly 76.76\%. This number is lower than that in POPE, where InstructBLIP achieved 88.73\% accuracy in the random setting \cite{li2023evaluating}. This is somewhat expected since attributes are finer details compared to object presence, and thus questions about attributes are harder than those about objects existence. In fact, the accuracy scores on object presence for LLaVa and InstructBLIP are approximately 10\% higher than their scores on attribute presence. On the other hand, mPLUG-Owl always performs worst, with its accuracy scores ranging between 50\% and 60\% across all settings. Surprisingly, this model performs slightly better on attributes than on object existence. In addition, the most challenging setting is our \textit{adversarial (image-based) setting}, where all the models suffer the worst accuracy scores. 

It is likely that the difference in performance between mPLUG-Owl and the other two models arises from mPLUG-Owl using LLAMA \cite{touvronLLaMAOpenEfficient} as its language model backbone, while the other two models use Vicuna \cite{vicuna2023}. The latter is fine-tuned on GPT-4 and ChatGPT conversations, while the former is not fine-tuned at all. While all three models fine-tune their final model in the multi-modal setting, it seems plausible that the fine-tuned language backbone provides improved instruction following ability, which makes LLaVa and InstructBLIP perform better on this benchmark.

\begin{table*}[t]
\centering
\begin{tabular}{llccccccccc}
\toprule
 &  & \multicolumn{3}{c}{Accuracy} & \multicolumn{3}{c}{F1} & \multicolumn{3}{c}{Yes (\%)} \\
 \cmidrule(lr){3-5} \cmidrule(lr){6-8} \cmidrule(lr){9-11}
 Sampling & Model & All & Att. & Obj. & All & Att. & Obj. & All & Att. & Obj. \\
\midrule
\multirow[c]{3}{*}{Random} & mPLUG-Owl & 56.04 & 57.21 & 53.81 & 60.35 & 55.08 & 67.16 & \textbf{62.00} & 45.00 & \textbf{93.00} \\
 & LLaVa & 71.36 & 66.62 & 80.85 & 72.55 & 66.85 & 82.84 & 54.00 & \textbf{51.00} & 62.00 \\
 & InstructBLIP & \textbf{76.76} & \textbf{72.22} & \textbf{85.83} & \textbf{74.96} & \textbf{67.95} & \textbf{86.51} & 43.00 & 37.00 & 55.00 \\
\cline{1-11}
\multirow[c]{3}{*}{Popular} & mPLUG-Owl & 62.63 & 66.32 & 55.51 & 64.61 & 61.58 & 68.27 & \textbf{56.00} & 38.00 & \textbf{92.00} \\
 & LLaVa & 76.47 & 72.41 & 84.60 & \textbf{75.64} & \textbf{69.69} & 85.70 & 47.00 & \textbf{41.00} & 58.00 \\
 & InstructBLIP & \textbf{76.75} & \textbf{72.50} & \textbf{85.23} & 74.58 & 67.42 & \textbf{86.00} & 41.00 & 34.00 & 56.00 \\
\cline{1-11}
\multirow[c]{3}{*}{Adv. Img.} & mPLUG-Owl & 56.19 & 57.14 & 54.33 & 59.92 & 53.92 & 67.63 & \textbf{60.00} & 43.00 & \textbf{92.00} \\
 & LLaVa & \textbf{68.26} & \textbf{64.82} & \textbf{75.12} & \textbf{69.46} & \textbf{64.02} & \textbf{78.61} & 54.00 & \textbf{48.00} & 66.00 \\
 & InstructBLIP & 67.48 & 64.05 & 74.31 & 66.81 & 59.76 & 77.70 & 48.00 & 39.00 & 65.00 \\
\cline{1-11}
\multirow[c]{3}{*}{Adv. Freq.} & mPLUG-Owl & 56.16 & 57.49 & 53.61 & 59.83 & 54.02 & 67.16 & \textbf{60.00} & 42.00 & \textbf{93.00} \\
 & LLaVa & 67.89 & 64.53 & 74.59 & \textbf{69.29} & \textbf{63.78} & 78.45 & 55.00 & \textbf{48.00} & 68.00 \\
 & InstructBLIP & \textbf{70.04} & \textbf{67.44} & \textbf{75.23} & 69.14 & 62.91 & \textbf{78.57} & 47.00 & 38.00 & 66.00 \\
\bottomrule
\end{tabular}
  \caption{Evaluation results on (All) all questions combined, (Att.) questions about attributes and (Obj.) questions about objects.}
  \label{tab:overall}
\end{table*}

\section{Further Analyses}
\subsection{Ablation 1: Positive Answer Bias}
POPE finds that the models' answers are highly skewed towards being positive (i.e., more likely to answer ``yes'') \cite{li2023evaluating}. Our results show a similar tendency (\Cref{tab:overall}), where a high overall ratio of ``yes'' to ``no'' is observed. However, this positive answer bias only holds for questions about object existence. For questions about attributes, models are more likely to answer ``no'' in most cases.

\subsection{Ablation 2: Context}
\label{sec:context}

\begin{figure}[h]
  \centering
  \includegraphics[width=0.5\textwidth]{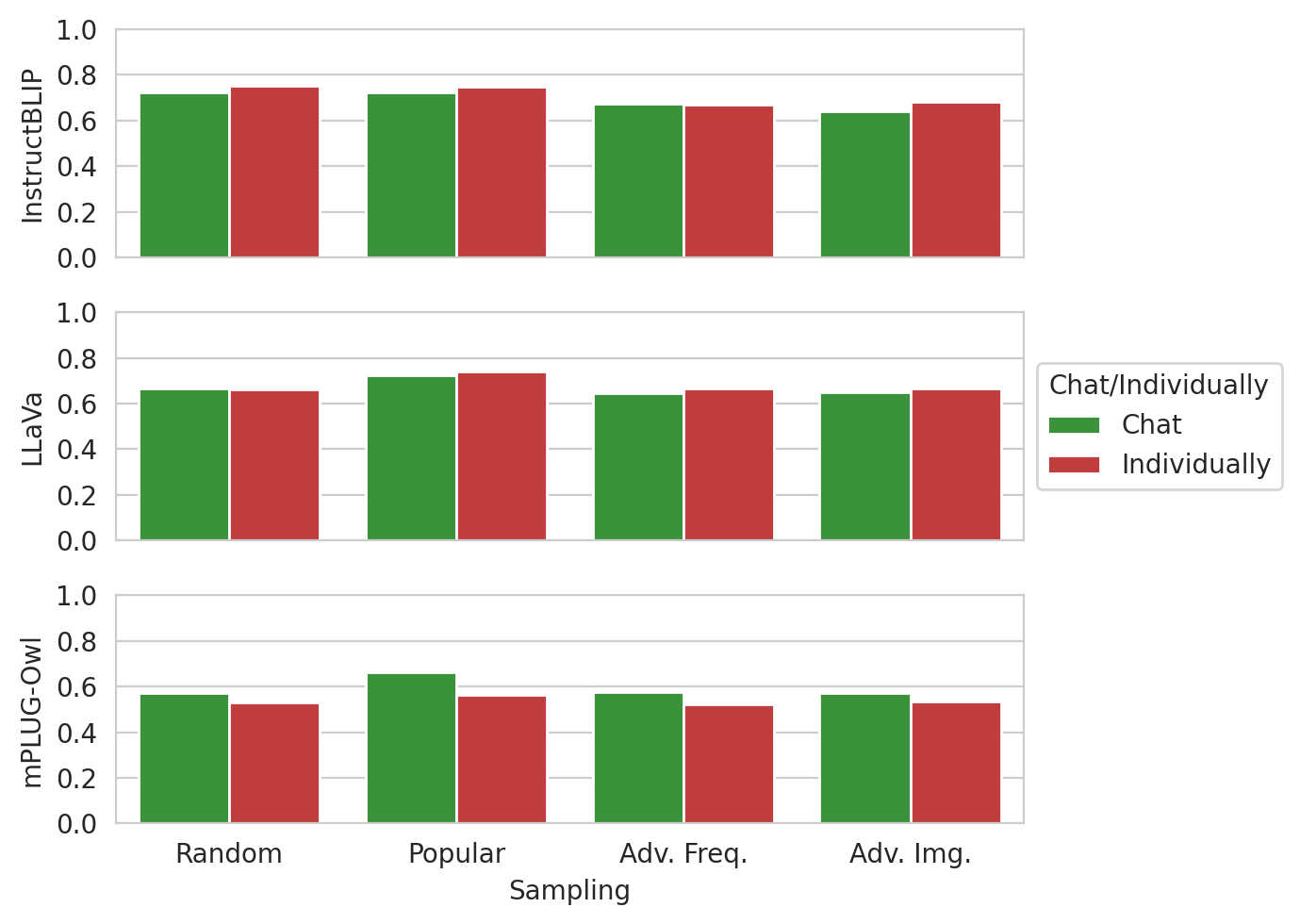}
  \caption{Difference in performance when asking our questions sequentially in a chat vs. asking
    them individually without context.}
  \label{fig:context}
\end{figure}

In our evaluation, questions are asked sequentially in a chat. We further investigate whether the models are sensitive to the context, which might contain incorrect answers in previous questions. An example of a chat (questions asked in context) is shown in \Cref{tab:qualex}. We therefore compare model performance with context vs. without context (\Cref{fig:context}). Similar results in both settings suggest that our benchmark remains stable regardless of the presence of context information. 
\begin{table*}[t]
  \centering
    \begin{tabular}{llll}
    \toprule
    Position & Question & Answer & Label \\
    \midrule
    1 & Is there a fence in the image? & Yes, there is a fence in the image. & yes \\
    2 & Is the fence of wooden material in the image? & No, the fence in the image is made of metal. & no \\
    3 & Is the fence of iron material in the image? & No, the fence in the image is made of metal. & yes \\
    4 & Is the fence of metal material in the image? & Yes, the fence in the image is made of metal. & yes \\
    5 & Is the fence black in the image? & Yes, the fence in the image is black. & no \\
    \bottomrule
    \end{tabular}
  \caption{Example of a chat given to the models.}
  \label{tab:qualex}
\end{table*}

\subsection{Visual Cue Analysis}
\begin{figure}[t]
  \centering
  \includegraphics[width=0.5\textwidth]{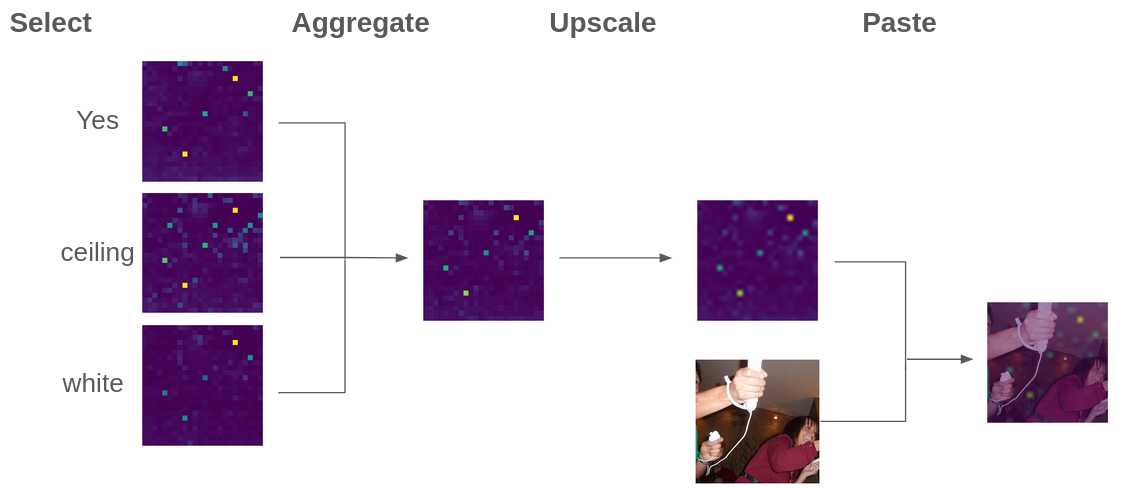}
  \caption{Pipeline for aggregating and visualizing the obtained relevance maps. We select the relevance maps for tokens that are either ``Yes'', ``No'' or name the object or attribute we asked about. The resulting maps are averaged. Afterwards, in line with \cite{stan2024lvlminterpretinterpretabilitytoollarge}, the maps are up-scaled to match the image size and plotted as a heatmap on top of the image.}
  \label{fig:relmappipeline}
\end{figure}

We study whether the models use the correct visual cues in the image to arrive at their answers, and thus gain insights into what causes hallucination. For example, in cases where the model does not hallucinate, one would expect the image patches corresponding to the object in question to be most relevant. 
We use LVLM-Interpret \cite{stan2024lvlminterpretinterpretabilitytoollarge} to visualize relevance maps \cite{cheferGenericAttentionmodelExplainability2021} for LLaVa \cite{llava} (figure \ref{fig:relmappipeline}). LVLM-Interpret returns one relevance map per predicted token in the answer, which indicates how
relevant each patch in the image was to predicting the corresponding token. To obtain a single
relevance map per answer, we select the relevance maps corresponding to the tokens \{``Yes'',
``No'', ``\$object'', ``\$attribute''\}, where \$object and \$attribute are replaced with the
object and attribute mentioned in the question. These maps are then averaged and upscaled to match the image size. To obtain the final visualization, they are plotted as heatmaps on top of the image. The last two steps follow the approach taken by LVLM-Interpret. 
In our qualitative examples (\Cref{fig:relmaps}), relevant image patches often correspond to the correct object. Interestingly, the locations of the relevant patches mostly remain the same, varying only in intensity. Due to this fact, no notable differences are observed between correct and hallucinated answers.

\newcommand\myawesomescale{0.15}
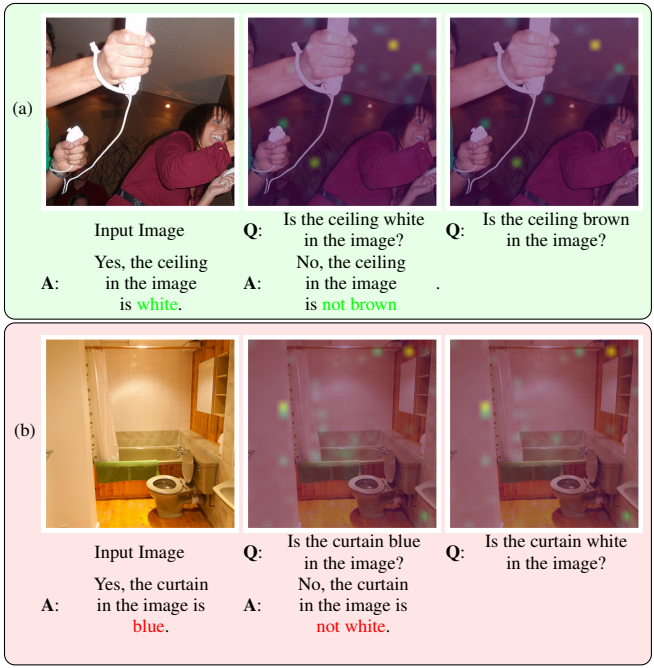
\begin{figure}[h]
  \scriptsize
  \noindent
  \centering
  	\begin{tikzpicture}[overlay]
		\draw [rounded corners,draw=black,fill=green!10] (-4.2,0.1) rectangle (4.4,-4.1);
		\draw [rounded corners,draw=black,fill=red!10] (-4.2,-4.15) rectangle (4.4,-8.7);
	\end{tikzpicture}
  \setlength{\tabcolsep}{1pt}
    \begin{tabular}{cccc}
    (a)
    & \includegraphics[valign=m, width=\myawesomescale\textwidth]{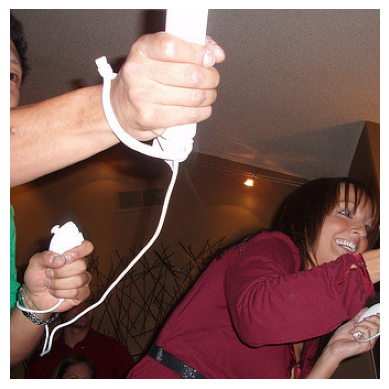}
    & \includegraphics[valign=m, width=\myawesomescale\textwidth]{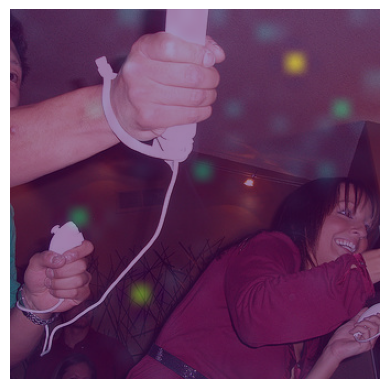}
    & \includegraphics[valign=m, width=\myawesomescale\textwidth]{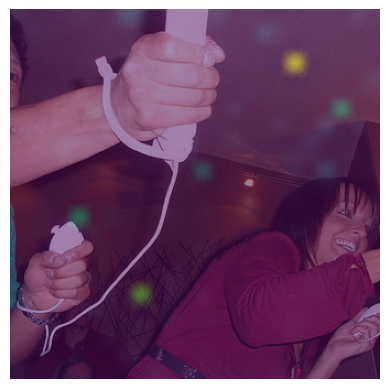}\\
    &  Input Image
    & \textbf{Q}: \makecell[c]{Is the ceiling white \\ in the image?}
    & \textbf{Q}: \makecell[c]{Is the ceiling brown \\ in the image?}\\
    & \textbf{A}: \makecell[c]{Yes, the ceiling \\ in the image \\ is {\color{green} white}.}
    & \textbf{A}: \makecell[c]{No, the ceiling \\ in the image \\ is {\color{green} not brown}}.\\
    \rule{0pt}{1.6cm} (b) 
    & \includegraphics[valign=m, width=\myawesomescale\textwidth]{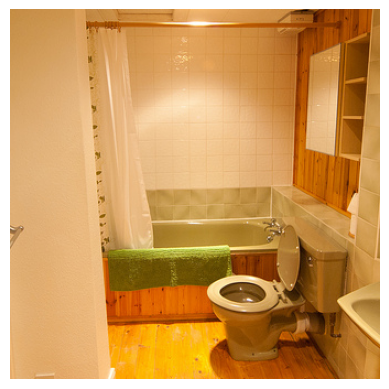}
    & \includegraphics[valign=m, width=\myawesomescale\textwidth]{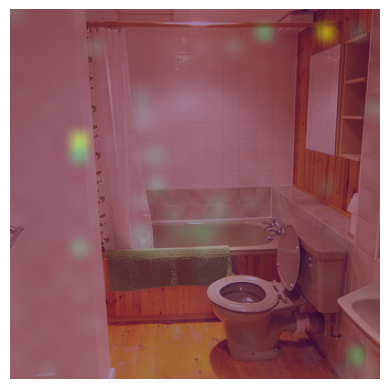}
    & \includegraphics[valign=m, width=\myawesomescale\textwidth]{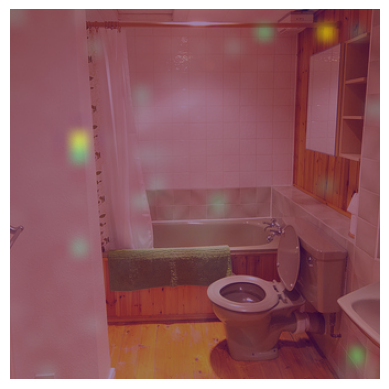}\\
    & Input Image
    & \textbf{Q}: \makecell[c]{Is the curtain blue \\ in the image?}
    & \textbf{Q}: \makecell[c]{Is the curtain white \\ in the image?}\\
    & \textbf{A}: \makecell[c]{Yes, the curtain \\ in the image is \\ {\color{red} blue}.}
    & \textbf{A}: \makecell[c]{No, the curtain \\ in the image is \\ {\color{red} not white}.}\\
  \end{tabular}
  \vspace{0.2cm}
  \caption{Examples of relevance maps for correct answers (in {\color{green}green}) and incorrect answers (in {\color{red}red}).}
  \label{fig:relmaps}
\end{figure}

\section{Limitations}
Our aim is to better understand hallucination on a finer-grained level, i.e. object attributes. Our H-POPE benchmark expects a clear yes or no answer from the model. While this is usually the case in our experiments, we also encounter situations where the models fail to do so. In specific, InstructBLIP missed 66 answers out of 91912, while LLaVa missed 58 answers, and mPLUG-Owl missed 417 answers. Nevertheless, the number of misses is very low compared to the total number of questions asked. Furthermore, our current setting is limited to only a few types of attributes, leaving other types such as tone, pattern, and orientation unexplored. Our H-POPE requires mutual exclusion in attributes, which makes it difficult to extend to some attribute types. We leave it to future work to devise more general sampling strategies.

\section{Conclusion}
In this work, we introduce H-POPE, an extension of POPE \cite{li2023evaluating}, to assess the tendency of LVLMs to hallucinate attributes of objects in images. Our benchmark demonstrates that current LVLMs hallucinate worse on attribute questions than on questions about object existence. In addition, our novel negative sampling strategy shows that these models are likely to assign attributes to the wrong objects in the image. Finally, analysing relevance maps does not provide insights into the source of hallucinations. We hope that this work will contribute to more accurate hallucination assessment of LVLMs.

\vspace{0.5cm}
\noindent \textbf{Acknowledgements}.
Our sincerest thanks goes to Haoran Wang for his valuable feedback throughout the development of the project.
The authors further thank the International Max Planck Institute for Informatics
(IMPRS-TRUST) for supporting Nhi Pham.
Michael Schott is supported by the Konrad Zuse School of Excellence in
Learning and Intelligent Systems (\href{https://eliza.school/}{ELIZA}) through the DAAD programme
Konrad Zuse Schools of Excellence in Artificial Intelligence, sponsored by the
Federal Ministry of Education and Research.

\clearpage
{\small
\bibliographystyle{ieee_fullname}
\bibliography{egbib}
}

\appendix

\section{Complete Evaluation Metrics}
We show complete tables including accuracy, f1-score, precision, recall and percentage of positive answers for all questions in \Cref{tab:overall_full}, for questions about objects in \Cref{tab:objects_full} and for questions about attributes in \Cref{tab:attributes_full}.

\begin{table*}[h]
  \centering
  \begin{tabular}{llrrrrr}
    \toprule
    &  & Accuracy & F1 & Precision & Recall & Yes (\%) \\
    Sampling & Model &  &  &  &  &  \\
    \midrule
    \multirow[c]{3}{*}{Random} & mPLUG-Owl & 56.04 & 60.35 & 54.28 & 67.95 & \textbf{62.00} \\
    & LLaVa & 71.36 & 72.55 & 69.68 & \textbf{75.66} & 54.00 \\
    & InstructBLIP & \textbf{76.76} & \textbf{74.96} & \textbf{81.18} & 69.63 & 43.00 \\
    \cline{1-7}
    \multirow[c]{3}{*}{Popular} & mPLUG-Owl & 62.63 & 64.61 & 60.74 & 69.00 & \textbf{56.00} \\
    & LLaVa & 76.47 & \textbf{75.64} & 78.39 & \textbf{73.08} & 47.00 \\
    & InstructBLIP & \textbf{76.75} & 74.58 & \textbf{82.20} & 68.25 & 41.00 \\
    \cline{1-7}
    \multirow[c]{3}{*}{Adv. Img.} & mPLUG-Owl & 56.19 & 59.92 & 54.83 & 66.04 & \textbf{60.00} \\
    & LLaVa & \textbf{68.26} & \textbf{69.46} & 66.91 & \textbf{72.21} & 54.00 \\
    & InstructBLIP & 67.48 & 66.81 & \textbf{68.15} & 65.52 & 48.00 \\
    \cline{1-7}
    \multirow[c]{3}{*}{Adv. Freq.} & mPLUG-Owl & 56.16 & 59.83 & 54.64 & 66.12 & \textbf{60.00} \\
    & LLaVa & 67.89 & \textbf{69.29} & 66.39 & \textbf{72.46} & 55.00 \\
    & InstructBLIP & \textbf{70.04} & 69.14 & \textbf{71.22} & 67.18 & 47.00 \\
    \bottomrule
  \end{tabular}
  \caption{Evaluation results on all questions combined.}
  \label{tab:overall_full}
\end{table*}

\begin{table*}[h]
  \centering
  \begin{tabular}{llrrrrr}
    \toprule
    &  & Accuracy & F1 & Precision & Recall & Yes (\%) \\
    Sampling & Model &  &  &  &  &  \\
    \midrule
    \multirow[c]{3}{*}{Random} & mPLUG-Owl & 53.81 & 67.16 & 50.85 & \textbf{98.90} & \textbf{93.00} \\
    & LLaVa & 80.85 & 82.84 & 75.10 & 92.35 & 62.00 \\
    & InstructBLIP & \textbf{85.83} & \textbf{86.51} & \textbf{82.36} & 91.10 & 55.00 \\
    \cline{1-7}
    \multirow[c]{3}{*}{Popular} & mPLUG-Owl & 55.51 & 68.27 & 52.09 & \textbf{99.03} & \textbf{92.00} \\
    & LLaVa & 84.60 & 85.70 & 79.95 & 92.35 & 58.00 \\
    & InstructBLIP & \textbf{85.23} & \textbf{86.00} & \textbf{81.60} & 90.91 & 56.00 \\
    \cline{1-7}
    \multirow[c]{3}{*}{Adv. Img.} & mPLUG-Owl & 54.33 & 67.63 & 51.68 & \textbf{97.84} & \textbf{92.00} \\
    & LLaVa & \textbf{75.12} & \textbf{78.61} & \textbf{68.91} & 91.47 & 66.00 \\
    & InstructBLIP & 74.31 & 77.70 & 68.49 & 89.78 & 65.00 \\
    \cline{1-7}
    \multirow[c]{3}{*}{Adv. Freq.} & mPLUG-Owl & 53.61 & 67.16 & 50.96 & \textbf{98.47} & \textbf{93.00} \\
    & LLaVa & 74.59 & 78.45 & 68.11 & 92.48 & 68.00 \\
    & InstructBLIP & \textbf{75.23} & \textbf{78.57} & \textbf{69.08} & 91.09 & 66.00 \\
    \bottomrule
  \end{tabular}
  \caption{Evaluation results on object questions.}
  \label{tab:objects_full}
\end{table*}

\begin{table*}[h]
  \centering
  \begin{tabular}{llrrrrr}
    \toprule
    &  & Accuracy & F1 & Precision & Recall & Yes (\%) \\
    Sampling & Model &  &  &  &  &  \\
    \midrule
    \multirow[c]{3}{*}{Random} & mPLUG-Owl & 57.21 & 55.08 & 57.96 & 52.48 & 45.00 \\
    & LLaVa & 66.62 & 66.85 & 66.39 & \textbf{67.32} & \textbf{51.00} \\
    & InstructBLIP & \textbf{72.22} & \textbf{67.95} & \textbf{80.29} & 58.90 & 37.00 \\
    \cline{1-7}
    \multirow[c]{3}{*}{Popular} & mPLUG-Owl & 66.32 & 61.58 & 71.66 & 53.98 & 38.00 \\
    & LLaVa & 72.41 & \textbf{69.69} & 77.30 & \textbf{63.44} & \textbf{41.00} \\
    & InstructBLIP & \textbf{72.50} & 67.42 & \textbf{82.69} & 56.92 & 34.00 \\
    \cline{1-7}
    \multirow[c]{3}{*}{Adv. Img.} & mPLUG-Owl & 57.14 & 53.92 & 58.31 & 50.14 & 43.00 \\
    & LLaVa & \textbf{64.82} & \textbf{64.02} & 65.51 & \textbf{62.59} & \textbf{48.00} \\
    & InstructBLIP & 64.05 & 59.76 & \textbf{67.87} & 53.38 & 39.00 \\
    \cline{1-7}
    \multirow[c]{3}{*}{Adv. Freq.} & mPLUG-Owl & 57.49 & 54.02 & 58.82 & 49.95 & 42.00 \\
    & LLaVa & 64.53 & \textbf{63.78} & 65.16 & \textbf{62.45} & \textbf{48.00} \\
    & InstructBLIP & \textbf{67.44} & 62.91 & \textbf{73.08} & 55.23 & 38.00 \\
    \bottomrule
  \end{tabular}
  \caption{Evaluation results on attribute questions.}
  \label{tab:attributes_full}
\end{table*}

We also show the prompt templates used for each model in \Cref{tab:prompts}. We followed the suggestions from the huggingface pages \footnote{\url{https://huggingface.co/docs/transformers/en/model_doc/llava}}\footnote{\url{https://huggingface.co/MAGAer13/mplug-owl-llama-7b}} of the corresponding models, with the exception of InstructBLIP, which did not provide one. We therefore use the default template suggested by huggingface \footnote{\url{https://huggingface.co/docs/transformers/en/chat_templating}}.

\lstset{
  basicstyle=\ttfamily,
  columns=fullflexible,
  keepspaces=true,
}

\begin{table*}
  \centering
  \begin{tabular}{rl}
    \toprule
    \multicolumn{1}{c}{Model} & \multicolumn{1}{c}{Template}\\
    \midrule
    mPLUG-Owl & \begin{lstlisting}[breaklines]
      Human: $question
      AI: $answer
      Human: $question
      AI: 
    \end{lstlisting} \\
    \cline{1-2}\\
    LLaVa & \begin{lstlisting}[breaklines]
      USER: $question ASSISTANT: $answer</s>USER: $question ASSISTANT: 
    \end{lstlisting} \\
    \cline{1-2}\\
    InstructBLIP & \begin{lstlisting}
      <|user|>
      $question
      <|assistant|>
      $answer
      <|user|>
      $question
      <|assistant|>
      
    \end{lstlisting} \\
    \bottomrule
  \end{tabular}
  \caption{Chat templates used for each model. The dollar sign \$ denotes that this is a variable to be
  replaced with either the question or the answer.}
  \label{tab:prompts}
\end{table*}

\end{document}